\documentclass{article}
\usepackage{ijcai23}

\usepackage{times}
\usepackage{soul}
\usepackage{url}
\usepackage[hidelinks]{hyperref}
\usepackage[utf8]{inputenc}
\usepackage[small]{caption}
\usepackage{graphicx}
\usepackage{booktabs}
\usepackage{algorithmic}
\usepackage[ruled]{algorithm2e}
\usepackage{extarrows}
\usepackage{subfigure}
\usepackage{amsmath}
\usepackage{amsmath,amssymb}
\usepackage{multicol}
\usepackage{mathrsfs} 
\usepackage{multirow}
\usepackage{color}
\usepackage[switch]{lineno}

\newtheorem{corollary}{Corollary}
\newtheorem{theorem}{Theorem}
\newtheorem{proof}{Proof}
\newtheorem{remark}{Remark}
\newtheorem{proposition}{Proposition}

\title{Few-shot Classification via Ensemble Learning with Multi-Order Statistics}

\author{
Sai Yang$^1$
\and
Fan Liu$^{2*}$\and
Delong Chen$^2$\And
Jun Zhou$^3$
\affiliations
$^1$School of Electrical Engineering, Nantong University, Nantong, China\\
$^2$College of Computer and Information, Hohai University, Nanjing, China\\
$^3$School of Information and Communication Technology, Griffith University, Queensland, Australia\\
\emails
$^{*}$Email: fanliu@hhu.edu.cn
}

\begin{document}

\maketitle

\begin{abstract}
    Transfer learning has been widely adopted for few-shot classification. Recent studies reveal that obtaining good generalization representation of images on novel classes is the key to improving the few-shot classification accuracy. To address this need, we prove theoretically that leveraging ensemble learning on the base classes can correspondingly reduce the true error in the novel classes. Following this principle, a novel method named Ensemble Learning with Multi-Order Statistics (ELMOS) is proposed in this paper. In this method, after the backbone network, we use multiple branches to create the individual learners in the ensemble learning, with the goal to reduce the storage cost. We then introduce different order statistics pooling in each branch to increase the diversity of the individual learners. The learners are optimized with supervised losses during the pre-training phase. After pre-training, features from different branches are concatenated for classifier evaluation. Extensive experiments demonstrate that each branch can complement the others and our method can produce a state-of-the-art performance on multiple few-shot classification benchmark datasets.
\end{abstract}

\section{Introduction}

\begin{figure}[h!]
		\includegraphics[width=1\linewidth]{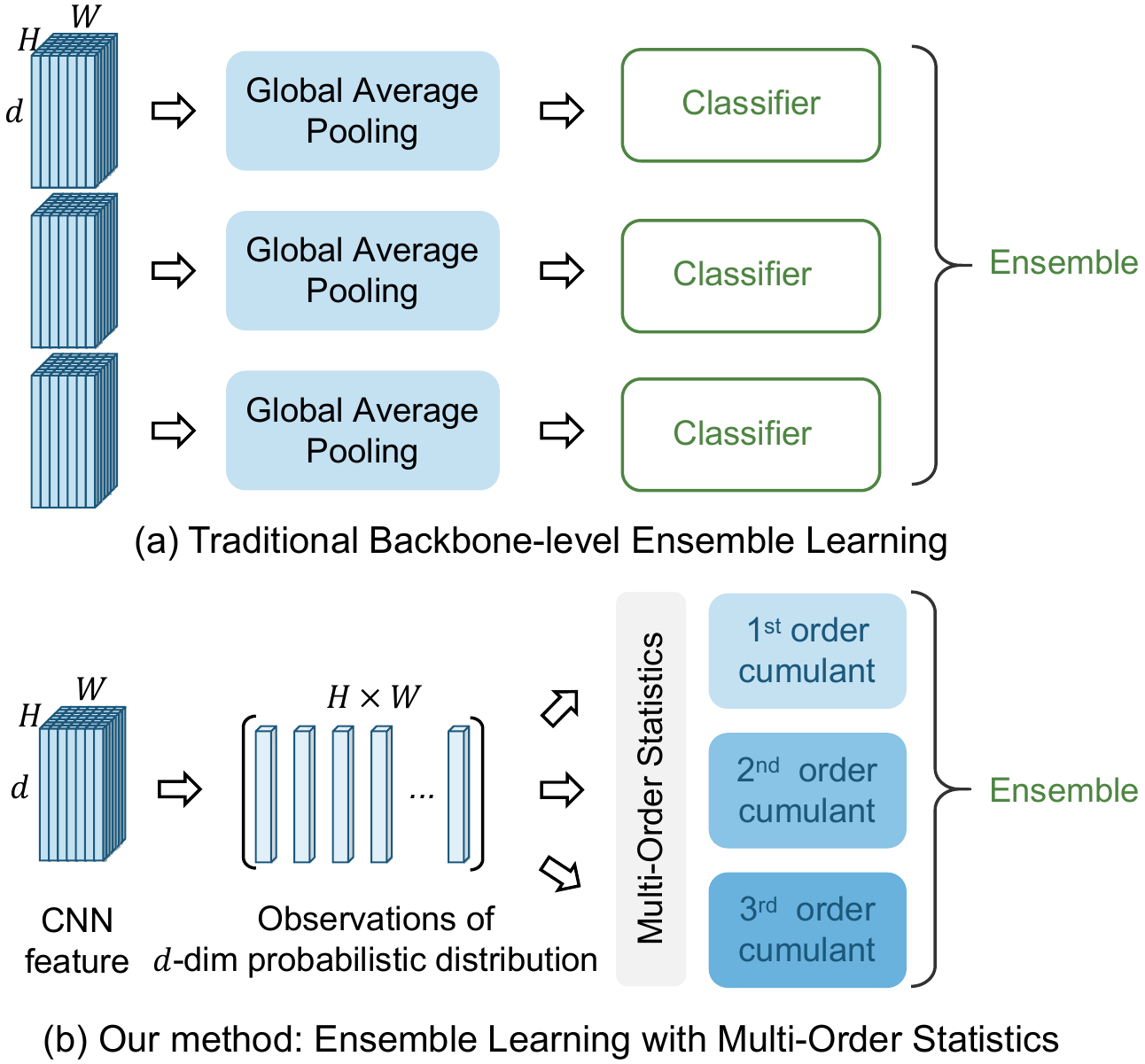}
		\caption{(a)~The traditional methods often use different backbone networks as individuals, which significantly increases the computation and storage costs. (b)~Our method takes the same backbone and equips different branches with multi-order statistics as learning individuals. They are parameter-free and trained jointly, and do not require extra model size and computation time.}
		\label{fig:motivation}
    \end{figure}

Few-shot Classification (FSC) is a promising direction in alleviating the labeling cost and bridging the gap between human intelligence and machine models. It aims to accurately differentiate novel classes with only a few labeled training samples. Due to limited supervision from novel classes, an extra base set with abundant labeled samples is often used to improve the classification performance. According to the adopted training paradigms,~FSC methods can be roughly divided into meta-learning-based ~\cite{finn2017model,snell2017prototypical} and transfer-learning-based~\cite{chen2019closer,liu2020negative,afrasiyabi2020associative}. The first type takes the form of episodic training, in which subsets of data are sampled from the base set to imitate the meta-test setting. Since sampling does not cover all combinations, this paradigm cannot fully utilize the information provided by the base set. In contrast, the transfer-learning takes the base set as a whole, so it avoids the drawback of meta-learning and achieves better performance. Many effective regularization techniques have been exploited in transfer-learning, for example, manifold mixup~\cite{mangla2020charting}, self-distillation~\cite{tian2020rethinking}, and self-supervised learning~\cite{zhang2020iept}, which leads to significant improvement on the generalization of image representations and the FSC performance.

Ensemble learning combines multiple learners to solve the same problem and exhibits better generalization performance than any individual learners~\cite{yang2013effective}. When combining ensemble learning with deep Convolutional Neural Networks (CNN), the new paradigm usually requires large-scale training data for classification tasks~\cite{horvath2021boosting,agarwal2021neural}, making it challenging to be adopted for FSC. Recently, two notable studies~\cite{dvornik2019diversity,bendou2022easy} employed an ensemble of deep neural networks for FSC tasks under either a meta-learning or a transfer-learning setting. They demonstrated that ensemble learning is also applicable to FSC. Yet, these works are still preliminary and lack a theoretical analysis to explain the underlying reason behind the promising performance. To address this challenge, we provide an FSC ensemble learning theorem for the transfer-learning regime. Its core idea is a tighter expected error bound on the novel classes, in which the expected error on the novel classes can be reduced by implementing ensemble learning on the base classes, given the base classes-novel classes domain divergence. 

The generalization ability of ensemble learning is strongly dependent on generating diverse individuals. As shown in Figure~\ref{fig:motivation} (a), traditional methods often use different backbone networks as individuals, which significantly increases the computation and storage costs. Our work finds that different-order statistics of the CNN features are complementary to each other, and integrating them can better model the whole feature distribution. Based on this observation, we develop a parameter-free ensemble method, which takes the same backbone and equips different branches with multi-order statistics as learning individuals. We name this method Ensemble Learning with Multi-Order Statistics (ELMOS), as shown in Figure~\ref{fig:motivation} (b). The main contributions of this paper are summarized as follows:
\begin{itemize}
    \item To our knowledge, this is the first theoretical analysis to guide ensemble learning in FSC. The derived theorem proves a tighter expected error bound is available on novel classes.
    \item We propose an ensemble learning method by adding multiple branches at the end of the backbone networks, which can significantly reduce the computation time of the training stage for FSC. 
    \item This is the first time that multi-order statistics is introduced to generate different individuals in ensemble learning.
    \item We conduct extensive experiments to validate the effectiveness of our method on multiple FSC benchmarks.
\end{itemize}

\section{Related Work}
In this section, we review the related work to the proposed method.
\subsection{Few-shot Classification}
    According to how the base set is used, FSC methods can be roughly categorized into two groups, meta-learning-based~\cite{zhang2020deepemd} and transfer-learning-based~\cite{chen2019closer,liu2020negative}. Meta-learning creates a set of episodes to simulate the real FSC test scenarios and simultaneously accumulate meta-knowledge for fast adaptation. Typical meta-knowledge includes optimization factors such as initialization parameters~\cite{finn2017model} and task-agnostic comparing ingredients of feature embedding and metric~\cite{snell2017prototypical,wertheimer2021few}. Recent literature on transfer learning~\cite{tian2020rethinking,chen2019closer} questioned the efficiency of the episodic training in meta-learning, and alternatively used all base samples to learn an off-the-shelf feature extractor and rebuilt a classifier for novel classes. Feature representations play an important role in this regime~\cite{tian2020rethinking}. To this end, regularization techniques such as negative-margin softmax loss and manifold mixup~\cite{liu2020negative,mangla2020charting} have been adopted to enhance the generalization ability of cross-entropy loss. Moreover, self-supervised~\cite{zhang2020iept,rajasegaran2020self} and self-distillation~\cite{ma2021partner,zhou2021binocular} methods have also shown promising performance in transfer-learning. To this end, supervised learning tasks can be assisted by several self-supervised proxy tasks such as rotation prediction and instance discrimination~\cite{zhang2020iept}, or by adding an auxiliary task of generating features during the pre-training~\cite{xu2021exploring}. When knowledge distillation is adopted, a high-quality backbone network can be evolved through multiple generations by a born-again strategy~\cite{rajasegaran2020self}. All these methods suggest the importance of obtaining generalization representations, and we will leverage ensemble learning to achieve this goal.

\subsection{Ensemble Learning}
    Ensemble learning builds several different individual learners based on the same training data and then combines them to improve the generalization ability of the learning system over any single learner. This learning scheme has shown promising performance on traditional classification tasks with deep learning on large-scale labeled datasets. Recently, ensemble learning for FSC methods has been presented. For example, \cite{dvornik2019diversity} combined an ensemble of prototypical networks through deep mutual learning under a meta-learning setting. \cite{bendou2022easy} reduced the capacity of each backbone in the ensemble and pre-trained them one by one with the same routine. However, the size of the ensemble learner increased for inference in the former work, while the latter required extra time to pre-train many learning individuals. Therefore, it still lacks efficient designs for learning individuals in FSC ensemble learning. Moreover, these works did not involve any theoretical analysis of the underlying mechanism of ensemble learning in FSC. In this paper, we investigate why ensemble learning works well in FSC under the transfer-learning setting. Based on the analysis, we propose an efficient learning method using a shared backbone network with multiple branches to generate learning individuals. 
    
\subsection{Pooling}
    Convolutional neural network models progressively learn high-level features through multiple convolution layers. A pooling layer is often added at the end of the network to output the final feature representation. To this end, Global Average Pooling (GAP) is the most popular option, however, it cannot fully exploit the merits of convolutional features because it only calculates the $1^{st}$-order feature statistics. Global Covariance Pooling (GCP) such as Deep$\rm O^2$P explores the $2^{nd}$-order statistic by normalizing the covariance matrix of the convolutional features, which has achieved impressive performance gains over the classical GAP in various computer vision tasks. Further research shows that using richer statistics may lead to further possible improvement. For example, Kernel Pooling~\cite{cui2017kernel} generates high-order feature representations in a compact form. However, a certain order statistic can only describe partial characteristics of the feature vector from the view of the characteristic function of random variables. For example, the first- and second-order statistics can completely represent their statistical characteristic only for the Gaussian distribution. Therefore, higher-order statistics are still needed for the non-Gaussian distributions, which are more ubiquitous in many real-world applications. This motivates us to calculate multi-order statistics to retain more information on features.

\section{The Proposed Method}
Here we present the proposed method. We start with a formal definition of FSC, and then present a theorem on FSC ensemble learning. This theorem leads to the development of an ensemble learning approach with multi-order statistics.

\subsection{Theory Foundation}
Under the standard setting of few-shot classification, three sets of data with disjoint labels are available, i.e., the base set $S_b$, the validation set $S_{val}$ and the novel set $S_{n}$. In the context of transfer-learning, $S_b$ is used for pre-training a model to well classify the novel classes in $S_{n}$, with the hyper-parameters tuned on $S_{val}$. Let $S_b=\{(x_i,y_i)\}_{i=1}^{N_b}$ denotes the source domain with ${N_b}$ labelled samples and $S_n$ denotes the target domain labelled with $K$ samples in each episode, where $N_b>>K$. Let the label function of $S_{b}$ and $S_{n}$ be $f_b$ and $f_n$, respectively. During the pre-training, a learner $h$ is obtained to approximate the optimal mapping function $h^*$ based on all ${N_b}$ training samples in $S_b$ from all possible hypotheses $\mathcal{H}$. When ensemble learning is introduced into the pre-training, several learners denoted as $\{h_o\}_{o=1}^{O}$ can be obtained. With the ensemble technique of weighted averaging, the final learner $\overline{h}$ is produced as: 
 \begin{equation}
    \label{eq:equ1}
    \overline{h}=\sum_{o=1}^{O}\alpha_oh_o,
    \end{equation}
where $\alpha_o$ is the weight parameter. There is a domain shift between the base and novel classes~\cite{tseng2020cross}, and we use the $L_1$ distance~\cite{kifer2004detecting} to measure the domain divergence between $S_{b}$ and $S_{n}$:
    \begin{equation}
    \label{eq:equ2}
    \mathcal{D}(S_b,S_n)=\int \left|\eta_{b}(x)-\eta_{n}(x)\right|\left|\overline{h}(x)-f_n(x)\right|dx,
    \end{equation}
where $\eta_b(x)$ and $\eta_n(x)$ is the density functions of $S_b$ and $S_n$ respectively.
\begin{theorem}[FSC Ensemble Learning]
    \label{thm-1}
        Let $\mathcal{H}$ be a hypothesis space, for any $h\in\{h_o\}_{o=1}^{O}\in \mathcal{H}$ is learned from $S_b$, and $\overline{h}=\sum_{o=1}^{O}\alpha_o h_o\in \mathcal{H}$, the expected error on $S_{n}$ respectively with $\overline{h}$ and $h$ holds the following relationship:
        \begin{gather*}
        e_n(\overline{h})\leq e_b(\overline{h})+\underbrace{\mathcal{D}(S_b,S_n)}_{\text{($S_b$-$S_n$) divergence}}+\lambda\\
       \leq e_b(h)+\underbrace{\mathcal{D}(S_b,S_n)}_{\text{($S_b$-$S_n$) divergence}}+\lambda,
        \end{gather*}
where $\lambda=E_{X\in{S_b}}\left|f_n(x)-f_b(x)\right|$ is a constant, $e_n(\overline{h})$ is the expected error on $S_n$ with $\overline{h}$, $e_b(h)$ is the expected error on $S_b$ with $h$, $e_b(\overline{h})$ is the expected error on $S_b$ with $\overline{h}$.
    \end{theorem}
    The proof is provided in the Supplementary Material.
\begin{remark}
The core idea of Theorem~\ref{thm-1} is to define a tighter expected error bound on the novel classes with the learned mapping function in the form of ensemble learning during the pre-training. Theorem~\ref{thm-1} tells that the true error on the novel classes can be reduced by implementing ensemble learning on the base classes, given the domain divergence between the novel class and base class. This can well explain the effectiveness of ensemble learning in few-shot classification, in which multiple learners are assembled to enhance the generalization on the base set, resulting in better performance in novel classes. 
\end{remark}
\subsection{FSC via Ensemble Learning with Multi-order Statistics}
\subsubsection{Overview}
\begin{figure*}[h!]
		\includegraphics[width=1\linewidth]{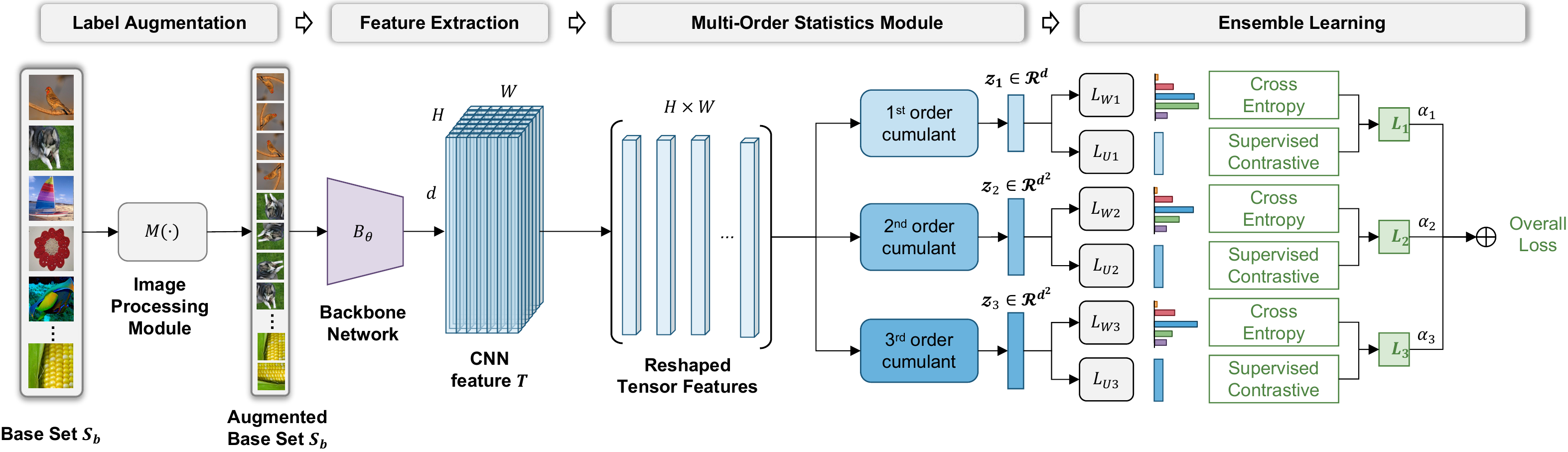}
		\caption{An overview of our framework. The images from $S_b$ are augmented by the image processing module and fed into the backbone for feature extraction. The CNN features from the backbone are then reshaped into the matrix, which is used to calculate multi-order statistics to equip different branches. Ensemble learning is implemented by the linear combination of multiple branches during the pre-training phase.}
		\label{fig:overview}
    \end{figure*}

Our method employs the transfer-learning paradigm in a two-phase manner. In the first phase, a good feature extractor is pre-trained on the base set. In the second phase, FSC evaluation is done on the novel set with the pre-trained feature extractor. Following Theorem~\ref{thm-1}, we introduce ensemble learning in the first phase to improve the FSC performance. The key to this phase is to effectively train multiple diverse individuals. Different from the previous works~\cite{dvornik2019diversity,bendou2022easy} that use many different networks as individuals, we add multiple branches after the backbone network to create individuals for reducing training costs. Each branch calculates different-order statistics for pooling to highlight the discrepancy between the individuals. This step is optimized by supervised losses. After pre-training, features from different branches are concatenated for FSC evaluation. We name this method as Ensemble Learning with multi-Order Statistics (ELMOS) for FSC. An overview of ELMOS is shown in Figure~\ref{fig:overview}, and a flow description of ELMOS is given in Algorithm~\ref{alg:algorithm1}. 

\subsubsection{Pre-training via Multi-order Statistics}
The proposed model architecture mainly consists of the following four components: an image processing module, the backbone network, a multi-order statistics module, and a supervised classifier module. The image processing module is denoted as $M\left(\cdot\right)$, which performs transformation of multi-scale rotation to augment the original base set and their label space. The backbone network is denoted as $B_\theta\left(\cdot\right)$ and parameterized by $\theta$, which converts each image into a tensor of size $ H \times W \times d$. The multi-order statistics module module is denoted as $S\left(\cdot\right)$, which maps the tensor from the backbone into multiple feature representations to generate individual learners for ensemble learning. The supervised classifier module is composed of softmax classifiers $L_W\left(\cdot\right)$ and the projectors $L_U\left(\cdot\right)$ with parameter matrices $W$ and $U$, respectively, which are used to build the supervised losses for pre-training. 

Given $L$ samples be randomly sampled from $S_{b}$ with $C_b$ classes, in which an image and its corresponding label are denoted as $(x_i, y_i)$, $y_i\in \{1,2,...C_b\}$. $M\left(\cdot\right)$ scales the images with the aspect-ratio of 2:3 and rotates the images with $\{0^\circ, 90^\circ, 180^\circ, 270^\circ \}$ under both the new and the original scales, resulting in eight times expansion of training samples. Feed $x_i$ into $B_\theta$ to produce a tensor feature of $T_i=B_\theta(x_i)\in \mathscr{R}^{ H \times W\times d}$. Next, we reshape the tensor $T_i$ into the matrix $T_i\in \mathscr{R}^{ HW\times d}$, and view each row vector in the matrix $t_j \in \mathscr{R}^{d}$ as an observation of the random variable of $t\in \mathscr{R}^{d}$. When $d=1$, the first characteristic function  of variable $t$ in the Laplace operator is given by:
 \begin{equation}
    \label{eq:equ3}
    \phi(s)=\int_ {-\infty}^{+\infty} f(t)e^{st}dt=\int_ {-\infty}^{+\infty} e^{st}dF(t),
\end{equation}
where $f(t)$ and $F(t)$ are the density function and distribution function of $t$, respectively. Let $\psi(s)=ln\phi(s)$ be the second characteristic function of the random variable $t$.
\begin{theorem}[The Inversion Formula for Distributions]
\label{thm:thm-2}
Let $t$ be a random variable with distribution function $F(t)$ and characteristic function $\phi(s)$. For $a,b\in C(F)$ and $a<b$,
\begin{gather*}
        F(b)-F(a)={\lim_{c \to \infty}}\frac{1}{2\pi}\int_{-c}^{c}\frac{e^{-sa}-e^{-sb}}{s}\phi(s)ds.
        \end{gather*}
\end{theorem}
 \begin{corollary}[Uniqueness] 
 \label{coro-1}
 If two distributions of $F_1(t)$ and $F_2(t)$ are identical, then the corresponding characteristic functions $\psi_1(s)$ and $\psi_2(s)$ are identical.
 \end{corollary}
See proof of Theorem~\ref{thm:thm-2} and Corollary~\ref{coro-1} in ~\cite{shiryaev2016probability}. From Theorem~\ref{thm:thm-2} and Corollary~\ref{coro-1}, we can see that there is a one-to-one correspondence between the characteristic function and the probability density function such that the characteristic function can completely describe a random variable.

The $o^{th}$-order cumulant of the random variable $t$ is defined as the $o^{th}$ derivative of function $\psi(s)$ at the origin, which is:
\begin{equation}
\label{eq:equ4}
    c_o=\frac{d^o\psi(s)}{ds^o}\bigg|_{s=0}.
\end{equation}
Then the Taylor series expansion of function $\psi(s)$ at the origin with respect to $s$ yields:
\begin{equation}
\label{eq:equ5}
    \psi(s)=c_1s+\frac{1}{2}c_2s^2+...+\frac{1}{o!}c_ps^o+R_s(s^o),
\end{equation}
where $R_s(s^o)$ is the remainder term. It can be seen from Equation~(\ref{eq:equ5}) that the $o^{th}$-order cumulant of $t$ is the coefficient of the term $s^o$ in Equation~(\ref{eq:equ5}). 
 \begin{proposition}
    \label{pro-1}
        Consider a Gaussion distribution $f(t)$ with mean $\mu$ and variance $\Sigma^2$ for the random variable $t$, its second characteristic function is:
        \begin{gather*}
       \psi(s)=\mu s+\frac{1}{2}\Sigma^2 s^2.
        \end{gather*}
    Consequently, the cumulant of the random variable $t$ are: 
      \begin{gather*}
       c_1=\mu, c_2=\Sigma^2, c_o=0\quad(o=3,4,...).
        \end{gather*}
\end{proposition}
The proof is provided in the Supplementary Material.
\begin{remark}
        Proposition~\ref{pro-1} implies that for Gaussian signals only, the cumulants are identically zero when the order is greater than 2. Please note this conclusion can be naturally extended to the scenario of multivariate variables when $d>1$. For the random variables with Gaussian distribution, the first and second-order statistics can completely represent their statistical characteristics. However, the non-Gaussian signals are more common in real-world applications. In this case, higher-order statistics also contain a lot of useful information. Therefore, we propose a multi-order statistics module consisting of multiple branches, each equipped with different order statistics of the tensor feature $T_i$.
\end{remark}

In particular, we employ three branches in the multi-order statistics module, which respectively calculate three orders cumulants of the variable $t$ with the observations in $T_i$. The specific formulation of the $1^{st}$-order, $2^{nd}$-order and $3^{rd}$-order cumulants of $t$ are expressed as:
\begin{equation}
\label{eq:equ6}
\begin{split}
    & c_{i1}=\frac{1}{H\times W}\sum_{j=1}^{H\times W}t_j\quad c_{i1}\in \mathscr{R}^{d},\\
    & c_{i2}=\frac{1}{H\times{W}}\sum_{j=1}^{H\times W}(t_j-c_{i1})(t_j-c_{i1})^T\quad c_{i2}\in \mathscr{R}^{d\times d},\\
    & c_{i3}=\frac{1}{H\times W}\sum_{j=1}^{H\times W}\frac{(t_j-c_{i1})^2(t_j-c_{i1})^T}{c_{i2}^2c_{i2}^T}\quad c_{i3}\in \mathscr{R}^{d\times d}.\\
\end{split}
\end{equation}
As $c_{i2}$ and $c_{i3}$ are $d\times d$ matrices, we flatten them into $d^2$-dimensional vectors and finally get the feature representations of $z_{i1}$, $z_{i2}$ and $z_{i3}$. We use these three features as individuals in ensemble learning, which respectively pass through their corresponding softmax classifier $L_W(\cdot)$ and projectors $L_U(\cdot)$. So the $o$-th ($o=1,2,3$) outputs are:
 \begin{equation}
   \label{eq:equ7}
    \begin{split}
        &P_{ij}^o=L_{Wo}\left(z_{io}\right)=\frac{exp({{z_{i0}}^Tw_{oj}})}{\sum_{j=1}^{8C_b}{exp({{z_{i0}}^Tw_{oj}})}},\\
        &{u_{io}}=\left\|{L_{Uo}}(z_{io})\right\|=\left\|{{z_{io}}^T}U_o\right\|,
     \end{split}
    \end{equation}
where $L_{Wo}(\cdot)$ is the $o$-th softmax classifier with the parameter matrix of $W_o$, $w_{oj}$ is the $j$-th component of $W_o$. $L_{Uo}(\cdot)$ is the $o$-th projector with the parameter matrix $U_o$. $P_{ij}^o$ is the $j$-th component of the output probability from the $o$-th softmax classifier. $u_{io}$ is the output vector from the $o$-th projector. We simultaneously employ Classification-Based (CB) loss of cross-entropy and Similarity-Based (SB) loss of supervised contrastive in supervised learning for each individual~\cite{scott2021mises}. These two losses are formulated as:
\begin{equation}
   \label{eq:equ8}
    \begin{aligned}
        &L_{CB}^{o}\left(\theta,W_o\right)=-{\sum_{i=1}^{8L}\sum_{j=1}^{8C_b}y_{ij}logP_{ij}^o},\\
        L_{SB}^{o}(\theta,U_o)=&-\sum_{i=1}^{8L}log\sum_{q\in{Q(u_{i0})}}\frac{exp(u_{io}\cdot u_{qo}/\tau)}{\sum_{a=1}^{8L}exp(u_{ao}\cdot u_{qo}/\tau)},
    \end{aligned}
    \end{equation}
where $y_{ij}$ is the $j$-th component of label $y_i$, $\tau$ is a scalar temperature parameter. $Q(u_{io})$ is the positive sample set, in which each sample has the same label as $u_{io}$. $u_{qo}$ is the $q$-th sample in $Q(u_{io})$. Then the learning objective function for the $o$-th individual is:
 \begin{equation}
 \label{eq:equ9}
   L_o(\theta,W_o,U_o)=L_{CB}^{o}\left(\theta,W_o\right)+ L_{SB}^{o}(\theta,U_o).
 \end{equation}
The overall loss function with ensemble learning is:
 \begin{equation}
 \label{eq:equ10}
   L_{overall}=\sum_{o=1}^{O}\alpha_o L_o (\theta,W_o,U_o),
 \end{equation}
where $\alpha_o$ is a weight controlling the contribution of each individual in the ensemble learning. The pre-training adopts the gradient descent method to optimize the above loss function. 
\begin{algorithm}[t]
\caption{Ensemble Learning with multi-Order Statistics (ELMOS) for FSC}
	\label{alg:algorithm1}
	\SetAlgoLined
	\KwIn {Base set $S_{b}$, support set $S_{p}$, query set $S_{q}$;
	augmentation module $M\left(\cdot\right)$, backbone network $B_\theta\left(\cdot\right)$, multi-order statistics module $S(\cdot)$, softmax classifier $L_{Wo}$, projector $L_{Uo}$ and logistic regression $g_\xi\left(\cdot\right)$;
	temperature parameter $\tau$, weight $\alpha_o$ ($o=1,2,3$). }
	\KwOut{Final prediction of the query samples}
	\textbf{Stage 1}: Pre-training  with ensemble learning
	
	\For{numbers of training epochs }{
    \textbf{Sample} a mini-batch with any image of $\{x_i,y_i\}$;
    
    \textbf{Feed} $x_i$ into $T\left(\cdot\right)$ and $B_\theta\left(\cdot\right)$ to obtain feature map $T_i\in \mathscr{R}^{ H \times W\times d}$ ;
    
    \textbf{Pass} $T_i$ through $S(\cdot)$ to output features $z_{io}$, ($o=1, 2, 3$);\
    
     \textbf{Pass} $z_{io}$ through $L_{Wo}$ and $L_{Uo}$ to get the output probability and projection feature ;

    \textbf{Calculate} optimization loss for each individual via Equation~(\ref{eq:equ9});
    
    \textbf{Calculate}  overall loss for pre-training via Equation~(\ref{eq:equ10});
    
    \textbf{Update} the parameters of $\theta$, $W_{o}$, $U_o$ using SGD;
   
    }
    
    \textbf{Stage 2}: Few-shot evaluation
    
    \For{all iteration = 1, 2, ..., MaxIteration}{
    \textbf{Feed}  $x_s\in {S_p}$ into $B_\theta(\cdot)$ and $S(\cdot)$ to output feature $z_{so}$, ($o=1,2,3$);
    
     \textbf{Concatenate} $z_{so}$ into the feature $z_s$ to train the classifier of $g_\xi\left(\cdot\right)$;
    }
    \textbf{Classify} the query samples according to Equation~(\ref{eq:equ13}).
\end{algorithm}

\subsubsection{Few-shot Evaluation}
The phase of few-shot evaluation still needs to construct a set of $N$-way $K$-shot FSC tasks, with a support set and a query set in each task. The support set randomly selects $K$ samples from each of the $N$ classes that are sampled from $S_n$, which is denoted as $S_{p}={\{x_s,y_s}\}_{s=1}^{NK}$, where $(x_s,y_s)$ is the $s$-th images and its corresponding label. The query set consists of the remaining images in these $N$ classes, which is denoted as $S_{q}=\{{x_q}\}_{q=1}^Q$ with any image of $x_q$. After pre-training, we get rid of the softmax classifier $L_W(\cdot)$ and projectors $L_U(\cdot)$ and fix the backbone network $B_\theta(\cdot)$ and the multi-order statistics module module $S(\cdot)$. The support set $S_{p}$ is input into $B_\theta(\cdot)$ and $S(\cdot)$ to produce the output features:
\begin{equation}
 \label{eq:equ11}
   z_{so}=B_\theta \circ S(x_s)\quad(o=1,2,3),
 \end{equation}
where $\circ$ is the stack operator. The features $z_{s1},z_{s2},z_{s3}$ are concatenated into a final expression of $x_s$:
\begin{equation}
 \label{eq:equ12}
   z_{s}= con( z_{s1}, z_{s2}, z_{s3}),
 \end{equation}
where $con(\cdot)$ is the concatenated operator. A logistic regression classifier $g_\xi\left(\cdot\right)$ parameterized by $\xi$ is then trained with $z_{s}$ and its corresponding label $y_s$. The query image $x_q$ is finally classified as:
\begin{equation}
\label{eq:equ13}
\hat{y}_q=g_\xi(z_q),
\end{equation}
where $\hat{y}_q$ is the inference label value of $x_q$.

\begin{table*}[t]
   \small 
    \caption{Test accuracy (\%) of each branch and their ensemble under 5-way 1-shot and 5-shot tasks on three datasets.}
    \label{tbl:Table1}
    \renewcommand\tabcolsep{10pt} %
    \begin{center}
    \begin{tabular}{cccccccc}
	\hline
	\multirow{2}{*}{Method} & \multirow{2}{*}{Backbone}& \multicolumn{2}{c}{miniImageNet} & \multicolumn{2}{c}{CIFAR-FS}& \multicolumn{2}{c}{CUB}  \\ \cline{3-8}
	&& 1-shot	& 5-shot	& 1-shot	& 5-shot& 1-shot	& 5-shot	\\\hline
    B\_1 & ResNet12& 69.06$\pm$0.44 & 83.61$\pm$0.29  & 77.09$\pm$0.46 & 88.46$\pm$0.34& 81.46$\pm$0.39 & 92.55$\pm$0.18 \\ 
	B\_2 & ResNet12& 66.42$\pm$0.42 & 85.76$\pm$0.26  & 71.53$\pm$0.48 & 88.83 $\pm$0.27& 77.79$\pm$0.39 & 94.44$\pm$0.17 \\ 
	B\_3 & ResNet12& 67.68$\pm$0.43 & 82.81$\pm$0.29  & 72.83$\pm$0.46 & 86.34$\pm$0.34& 83.89$\pm$0.38 & 91.20$\pm$0.17 \\ 
   ELMOS& ResNet12& 70.30$\pm$0.45 & 86.17$\pm$0.26 & 78.18$\pm$0.41 & 89.87$\pm$0.31& 85.21$\pm$0.38 & 95.02$\pm$0.16 \\ 
	\hline
    \end{tabular}
    \end{center}
\end{table*}
\begin{figure}[t]
\hspace{-0.2cm}
	\subfigure[5-way 1-shot]{
	\begin{minipage}[t]{0.5\linewidth}
    \includegraphics[width=1.7in]{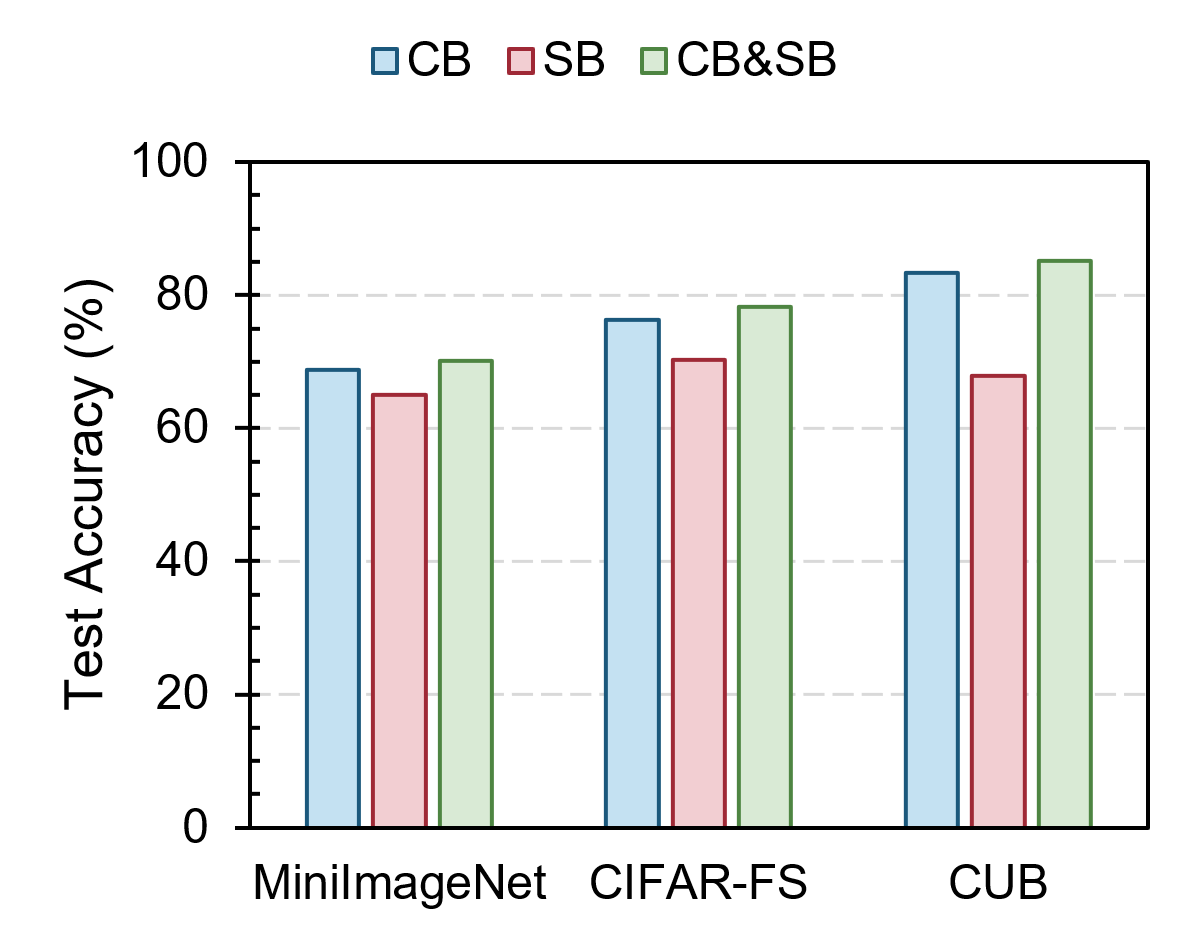}
    \end{minipage}%
     }%
    \subfigure[5-way 5-shot]{
	\begin{minipage}[t]{0.5\linewidth}
    \includegraphics[width=1.7in]{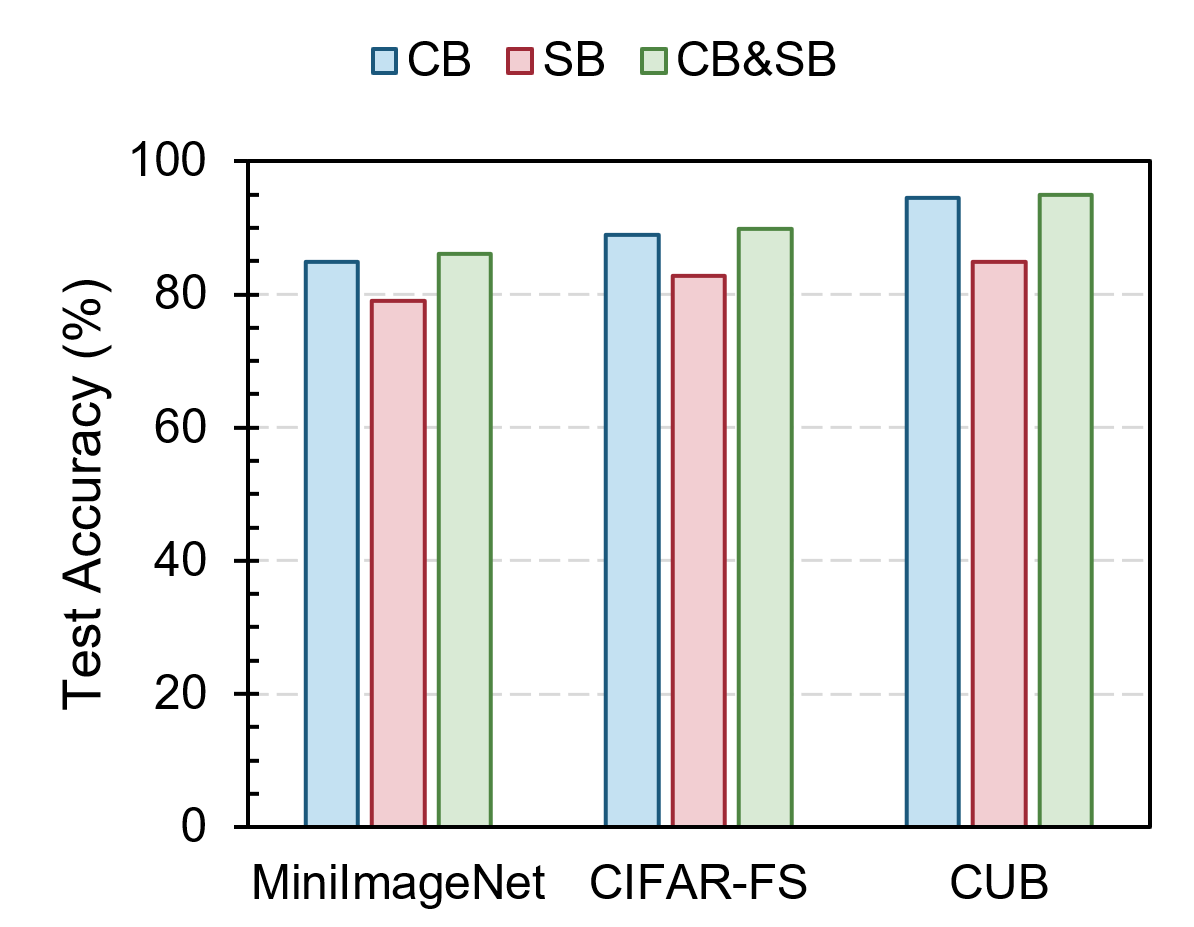}
    \end{minipage}%
     }%
\caption{Test accuracy (\%) of the classification-based (CB) loss, similarity-based (SB) loss and their combination (CB\&SB) under 5-way 1-shot and 5-way 5-shot tasks on three datasets.}
\label{fig:Loss}
\end{figure}
\begin{table}[ht]
    \small
     \renewcommand\tabcolsep{2.5pt} 
    \caption{Comparison of results against state-of-the-art methods on CUB dataset.The top three results are marked in red, blue and green.}
    \label{tbl:Table2}
    \begin{center}
    \begin{tabular}{p{4.1cm}ccc}
	\hline
	\multirow{2}{*}{Method}  & \multicolumn{2}{c}{CUB}
	\\ \cline{2-3} &1-shot &5-shot  \\ \hline
	\textbf{Meta-learning} \\
	Relational \cite{sung2018learning} &55.00$\pm$1.00 &69.30$\pm$0.80 \\
	DeepEMD \cite{zhang2020deepemd}&75.65$\pm$0.83&88.69$\pm$0.50\\
	BML \cite{zhou2021binocular}&76.21$\pm$0.63&90.45$\pm$0.36 \\ 
	RENet \cite{kang2021relational}&79.49$\pm$0.44&91.11$\pm$ 0.24\\
	FPN\cite{wertheimer2021few}&\textcolor{green}{83.55$\pm$0.19} &\textcolor{green}{92.92$\pm$0.10}\\
	IEPT \cite{zhang2020iept}&69.97$\pm$0.49&84.33$\pm$0.33 \\
	APP2S \cite{ma2022adaptive}&77.64$\pm$0.19&90.43$\pm$0.18   \\
    MFS \cite{afrasiyabi2022matching} &79.60$\pm$0.80&90.48$\pm$0.44 \\
    DeepBDC \cite{xie2022joint}&\textcolor{blue}{84.01$\pm$0.42}&\textcolor{blue}{94.02$\pm$0.24}  \\
    HGNN \cite{yu2022hybrid}& 78.58$\pm$0.20&90.02$\pm$0.12 \\
    INSTA\cite{Ma2022Learning}&75.26 ± 0.31&88.12 ± 0.54\\
	\textbf{Transfer-learning} \\
	Baseline++ \cite{chen2019closer}&60.53$\pm$0.83&79.34$\pm$0.61\\
	Neg-Cosine \cite{liu2020negative}&72.66$\pm$0.85&89.40$\pm$0.43\\
	S2M2 \cite{mangla2020charting} &80.68$\pm$0.81&90.85$\pm$0.44 \\
	DC-LR\cite{yang2021free}&79.56$\pm$0.87&90.67$\pm$0.35\\
	CCF \cite{xu2021exploring}&81.85$\pm$0.42&91.58$\pm$0.32\\
	ELMOS (ours)&\textcolor{red}{85.21$\pm$0.38} &\textcolor{red}{95.02$\pm$0.16}  \\
	\hline
    \end{tabular}

    \end{center}
\end{table}
\section{Experiments}
\subsection{Datasets}
\textbf{miniImageNet} contains 600 images over 100 classes, which are divided into 64, 16 and 20 respectively for base, validation and novel sets.~\textbf{tiredImageNet} consists of 779, 165 images belonging to 608 classes, which are further grouped into 34 higher-level categories with 10 to 30 classes per category. These categories are partitioned into 20 categories (351 classes), 6 categories (97 classes) and 8 categories (160 classes) respectively for base, validation and novel sets.~\textbf{CIFAR-FS} is derived from CIFAR100 and consists of 100 classes with 600 images per class. The total classes are split into 64, 16 and 20 for base, validation and novel sets.~\textbf{Caltech-UCSD Bird-200-2011(CUB)} has a total number of 11,788 images over 200 bird species. These species are divided into 100, 50, and 50 for the base, validation and novel sets, respectively.

\subsection{Implementation Details}
In the experiments, we primarily used ResNet12 architecture with 4 residual blocks. Each block had 3 convolutional layers with 3×3 kernels. The number of kernels for the 4 blocks was 64, 160, 320, and 640, respectively. A max-pooling layer was added at the end of the first three blocks. The last block was branched with three pooling layers, which respectively modeled different statistical representations of the images. We opted for the SGD optimizer with a momentum of 0.9 and a weight decay of 5e-4. The learning rate was initialized to be 0.025. We trained the network for 130 epochs with a batch size of 32 in all the experiments. For miniImageNet, tiredImageNet and CIFAR-FS, the learning rate was reduced by a factor of 0.2 at the 70-$th$ and 100-$th$ epoch. For CUB, the learning rate was reduced by a factor of 0.2 for every 15 epochs after the 75-$th$ epoch. We randomly sampled 2,000 episodes from $S_{n}$ with 15 query samples per class for both 5-way 1-shot and 5-shot evaluations, to produce the mean classification accuracy as well as the 95\% confidence interval.

\begin{table*}[ht]
    \renewcommand\tabcolsep{3pt} 
    \small
    \caption{Comparison of results against state-of-the-art methods on miniImageNet, tiredImageNet, and CIFAR-FS dataset. '-' means the results were not provided by the authors. The top three results are marked in red, blue and green, respectively.}
    \label{tbl:Table3}
    \begin{center}
    \begin{tabular}{p{4.2cm}ccccccccc}
	\hline
	\multirow{2}{*}{Method}  &\multirow{2}{*}{Backbone}&\multirow{2}{*}{Venue} & \multicolumn{2}{c}{miniImageNet}& \multicolumn{2}{c}{tiredImageNet}& \multicolumn{2}{c}{CIFAR-FS}
	\\ \cline{4-9}  &&&1-shot &5-shot &1-shot &5-shot&1-shot &5-shot \\ \hline
	\textbf{Meta-learning} \\
    DeepEMD \cite{zhang2020deepemd}& ResNet12 &CVPR'20&65.91$\pm$0.82 &82.41$\pm$ 0.56 &71.16$\pm$0.87&86.03$\pm$0.58&-&- \\
    CC+rot \cite{gidaris2019boosting}&ResNet12&CVPR’20&62.93$\pm$0.45&79.87$\pm$0.33&70.53$\pm$0.51 &84.98$\pm0.36$ &76.09$\pm$0.30&87.83$\pm$0.21\\ 
	BML \cite{zhou2021binocular}&ResNet12&ICCV’21&67.04$\pm$0.63&83.63$\pm$0.29&68.99$\pm$0.50&85.49$\pm$0.34&73.45$\pm$0.47&88.04$\pm$0.33\\ 
	RENet \cite{kang2021relational} &ResNet12&ICCV’21&67.60$\pm$0.44&82.58$\pm$0.30&71.61$\pm$0.51&85.28$\pm$0.35&74.51$\pm$0.46&86.60$\pm$0.32   \\
    MeTAL\cite{baik2021meta}&ResNet12&CVPR’21&66.61$\pm$0.28&81.43$\pm$0.25&70.29$\pm$0.40&86.17$\pm$0.35&-&-\\
	DAN \cite{xu2021learning}&ResNet12&CVPR’21&67.76$\pm$0.46&82.71$\pm$0.31&71.89$\pm$0.52 &85.96$\pm$0.35&-&-\\
	IEPT \cite{zhang2020iept} &ResNet12&ICLR'21 &67.05$\pm$0.44 &82.90$\pm$0.30 &72.24$\pm$0.50&86.73$\pm$0.34&-&-  \\
	APP2S \cite{ma2022adaptive}&ResNet12&AAAI’22&66.25$\pm$0.20&83.42$\pm$0.15&72.00$\pm$0.22 &86.23$\pm$0.15&73.12 $\pm$0.22&85.69$\pm$0.16   \\
   DeepBDC \cite{xie2022joint}&ResNet12&CVPR’22&67.34$\pm$0.43&84.46$\pm$0.28&\textcolor{green}{72.34$\pm$0.49}&\textcolor{green}{87.31$\pm$0.32}&-&- \\
   MFS \cite{afrasiyabi2022matching}&ResNet12&CVPR’22&68.32$\pm$0.62&82.71$\pm$0.46&\textcolor{blue}{73.63$\pm$0.88}&\textcolor{blue}{87.59$\pm$0.57}&-&- \\
   TPMN\cite{wu2021task}&ResNet12&CVPR’22&67.64$\pm$0.63&83.44$\pm$0.43&72.24$\pm$ 0.70 &86.55 $\pm$ 0.63&-&-\\
   HGNN \cite{yu2022hybrid}&ResNet12&AAAI’22&67.02$\pm$0.20&83.00$\pm$0.13&72.05$\pm$0.23& 86.49$\pm$0.15&-&-\\
    DSFN\cite{Zhang2022kernel}& ResNet12&ECCV'22&61.27$\pm$0.71& 80.13$\pm$0.17&65.46$\pm$ 0.70&82.41$\pm$0.53&-&-\\
    MTR\cite{Bouniot2022Improving}& ResNet12&ECCV'22 &62.69$\pm$ 0.20 &80.95$\pm$0.14& 68.44 $\pm$0.23& 84.20 $\pm$0.16&-&-\\
	\textbf{Transfer-learning} \\
	Baseline++ \cite{chen2019closer}&ResNet12 &ICLR'19 &48.24$\pm$0.75 &66.43$\pm$0.63 &-&--&-\\
	Neg-Cosine \cite{liu2020negative}&WRN28&ECCV'20 &61.72$\pm$0.81 &81.79$\pm$0.55 &-&--&-\\
	RFS \cite{tian2020rethinking}&WRN28 &ECCV'20 &64.82$\pm$0.60 &82.14$\pm$0.43&71.52$\pm$0.69&86.03$\pm$0.49&-&- \\
	CBM \cite{wang2020cooperative} &ResNet12&MM'20 &64.77$\pm$0.46 &80.50$\pm$0.33 &71.27$\pm$0.50&85.81$\pm$0.34&-&-\\
	SKD \cite{rajasegaran2020self}&ResNet12&Arxiv’21 &67.04$\pm$0.85 &83.54$\pm$0.54&72.03$\pm$0.91&86.50$\pm$0.58&76.9$\pm$0.9&\textcolor{green}{88.9$\pm$0.6} \\
	IE\cite{rizve2021exploring}&ResNet12 &CVPR’21 &67.28$\pm$0.80 &\textcolor{blue}{84.78$\pm$0.33} &72.21$\pm$0.90&87.08$\pm$0.58&\textcolor{blue}{77.87$\pm$0.85}&\textcolor{blue}{89.74$\pm$0.57}\\
	PAL \cite{ma2021partner}&ResNet12 &ICCV’21 &\textcolor{blue}{69.37$\pm$0.64}&84.40$\pm$0.44&72.25$\pm$0.72&86.95$\pm$0.47&\textcolor{green}{77.1$\pm$0.7}&88.0$\pm$0.5\\
	CCF\cite{xu2021exploring}&ResNet12&CVPR’22&\textcolor{green}{68.88$\pm$0.43}&\textcolor{green}{84.59$\pm$0.30}	&-	&-&-	&-\\
	ELMOS (ours) &ResNet12&- &\textcolor{red}{70.30$\pm$0.45} &\textcolor{red}{86.17$\pm$0.26} &\textcolor{red}{73.84$\pm$0.49 }&\textcolor{red}{87.98$\pm$0.31}&\textcolor{red}{78.18$\pm$0.41 }&\textcolor{red}{89.87$\pm$0.31} \\
	\hline
    \end{tabular}
\end{center}
\end{table*}
\begin{table}[ht]
     \renewcommand\tabcolsep{2.5pt}
    \small
    \caption{Comparison of results with the most related method under 5-way 1-shot and 5-shot tasks on CIFAR-FS and CUB.}
    \label{tbl:Table5}
    \begin{center}
   \begin{tabular}{p{1cm}ccccc}
	\hline
	\multirow{2}{*}{Method} & \multicolumn{2}{c}{CIFAR-FS}& \multicolumn{2}{c}{CUB}& 
	\\\cline{2-5}  &1-shot &5-shot &1-shot &5-shot\\ \hline
   EASY & 75.24$\pm$0.20 &88.38$\pm$0.14 & 77.97$\pm$0.20 &91.59$\pm$0.10\\
   ELMOS&78.18$\pm$0.41&89.87$\pm$0.31&85.21$\pm$0.38 &95.02$\pm$0.16\\ 
\hline
    \end{tabular}
\end{center}
\end{table}
\subsection{Ablation Studies}
The effectiveness of our method is attributed to the ensemble of different branches equipped with multi-order statistics. In this section, we conducted ablation studies to analyze the effect of the $1^{st}$-order, $2^{nd}$-order and, $3^{rd}$-order statistical pooling and their combination on the miniImageNet, CIFAR-FS and CUB datasets. Above methods are respectively denoted as B\_1, B\_2, B\_3,and ELMOS. Their accuracies under 5-way 1-shot and 5-shot tasks on three datasets are shown in Table~\ref{tbl:Table1}. From the results, we can see that: (1) On all three datasets, the test accuracy of $B_1$ and $B_3$ is higher than $B_2$ under the 1-shot task, but the test accuracy of $B_2$ is higher than $B_1$ and $B_3$ under the 5-shot task. The above phenomenon shows that different order statistics provide different information about the images. (2) The test accuracy of ELMOS is higher than $B_1$, $B_2$ and $B_3$ under both 1-shot and 5-shot tasks, which illustrates that different order statistics complement each other. Combing them can bring more useful information for classification, resulting in higher classification performance. 

For each individual in the ensemble learning, the optimization is cooperatively accomplished by the Classification-Based (CB) loss and Similarity-Based (SB) loss~\cite{scott2021mises}. Hence, we conducted ablation experiments to analyze the contribution of each loss on three benchmark datasets: miniImageNet, CIFAR-FS and CUB. Subsequently, we pre-trained the model respectively with CB and SB loss alone and their combination, resulting in three methods denoted as CB, SB and CB\&SB. The test accuracies under different methods are shown in Figure~\ref{fig:Loss}. The test results show that the accuracy of CB\&SB is higher than CB and SB, which implies that both classification-based and similarity-based losses play important roles in our method.
\subsection{Comparison with the Most Related Method}
Our method is most related to EASY~\cite{bendou2022easy}, which is also a FSC ensemble learning method in context of transfer learning. The comparison of results between them on CIFAR-FS and CUB datasets is shown in Table~\ref{tbl:Table5}. From the results, we can see that our method beats EASY by a very large margin under both 1-shot and 5-shot tasks. Please note that our method is more efficient that EASY, because EASY needs to pre-train multiple individual networks, which spends much more pre-training time than our method.

\subsection{Comparison with State-of-the-Art Methods}
We compare the performance of our method with several state-of-the-art methods. These methods are either meta-learning based or transfer-learning based. The comparison of results is shown in Table~\ref{tbl:Table2} and Table~\ref{tbl:Table3}. From Table~\ref{tbl:Table2}, we can see the performance of our method ranks at the top under both 1-shot and 5-shot tasks on CUB. Specifically, our method exceeds the second-best model DeepBDC by 1.2\% and 1.0\% respectively in 1-shot and 5-shot settings. From Table~\ref{tbl:Table3}, we can see that our method beats state-of-the-art methods under both 5-way 1-shot and 5-way 5-shot tasks on the dataset of miniImageNet, tiredImagegNet, and CIFAR-FS. Specifically, on miniImageNet, PAL and IE behave the second best respectively in 1-shot and 5-shot settings. Our method beats them by 0.93\% and 1.39\%. On tiredImageNet, our method outperforms the second-best MFS by 0.21\% and 0.39\% respectively in 1-shot and 5-shot settings. On CIFAR-FS, our method achieves 0.31\% and 0.13\% improvement over IE for 1-shot and 5-shot respectively. In brief, our method consistently outperforms the state-of-the-art FSC methods under both 1-shot and 5-shot tasks on multiple datasets. The promising results are achieved because of the generalization representation obtained by ensemble learning with multi-order on the base set.
\section{Conclusion}
This paper analyzes the underlying work mechanism of ensemble learning in few-shot classification. A theorem is provided to illustrate that the true error on the novel classes can be reduced with ensemble learning on the base set, given the domain divergence between the base and the novel classes. Multi-order statistics on image features are further introduced to produce learning individuals to get an effective ensemble learning design. Comprehensive experiments on multiple benchmarks have illustrated that different-order statistics can generate diverse learning individuals due to their complementarity. The promising FSC performance with ensemble learning on the base set has validated the proposed theorem. 
\bibliographystyle{named}
\bibliography{ijcai23}
\section{Supplementary Material}
\subsection{Proof of Theorem 1}
\begin{proof}
 The expected error with $\overline{h}$ on $S_{b}$ and $S_{n}$ are:
\begin{equation}
    \label{eq:equ14}
    \begin{aligned}
      &e_b(\overline{h})=e_{S_b}(\overline{h},f_b)=E_{x\in S_b}[\left|\overline{h}(x)-f_b(x)\right|]\\
      &e_n(\overline{h})=e_{S_n}(\overline{h},f_n)=E_{x\in S_n}[\left|\overline{h}(x)-f_n(x)\right|].
       \end{aligned}
    \end{equation}
Then, we will get the following formula:
\begin{equation}
    \label{eq:equ15}
    \begin{aligned}
      &e_n(\overline{h})=e_n(\overline{h})+e_b(\overline{h})-e_b(\overline{h})+e_{S_b}(\overline{h},f_n)-e_{S_b}(\overline{h},f_n)\\
      &\leq e_b(\overline{h})+\left|e_{S_b}(\overline{h},f_n)-e_{S_b}(\overline{h},f_b)\right|+\\
      &\left|e_{S_{n}}(\overline{h},f_n)-e_{S_b}(\overline{h},f_n)\right|\\
      &\leq e_b(\overline{h})+E_{X\in{S_b}}\left|f_n(x)-f_b(x)\right|+\\
      &\left|e_{S_{n}}(\overline{h},f_n)-e_{S_b}(\overline{h},f_n)\right|\\
&\leq e_b(\overline{h})+E_{X\in{S_b}}\left|f_n(x)-f_b(x)\right|+\\
&\int \left|\eta_{b}(x)-\eta_{n}(x)\right|\left|\overline{h}(x)-f_n(x)\right|dx\\
&\leq e_b(\overline{h})+E_{X\in{S_b}}\left|f_n(x)-f_b(x)\right|+\mathcal{L}(S_b,S_n).
       \end{aligned}
    \end{equation}
The expected error on $S_b$ with any learner $h_o$ of ensemble learning is calculated as:
\begin{equation}
    \label{eq:equ16}
    \begin{aligned}
      e_b(h_o)=\int (h_o(x)-h^*(x))^2 \eta_b(x)dx,
       \end{aligned}
    \end{equation}
where $\eta_b(x)$ is the density functions of $S_b$, The average error on $S_b$ with the learners of ensemble learning is:
\begin{equation}
    \label{eq:equ17}
    \begin{aligned}
      e_b(h)= \sum_{o=1}^{O}\alpha_o\int (h_o(x)-h^*(x))^2 \eta_b(x)dx.
       \end{aligned}
    \end{equation}
Recall that $\overline{h}=\alpha_o\sum_{o=1}^{O}h_o$, then the expected error on $S_b$ with $\overline{h}$ is calculated as:
\begin{equation}
    \label{eq:equ18}
    \begin{aligned}
      &e_b(\overline{h})=\int (\overline{h}(x)-h^*(x))^2 \eta_b(x)dx\\
      &=\int(\alpha_o\sum_{o=1}^{O}h_o(x)-h^*(x))^2 \eta_b(x)dx\\
      &\leq\sum_{o=1}^{O}\alpha_o\int (h_o(x)-h^*(x))^2 \eta_b(x)dx\leq e_b(h).
       \end{aligned}
    \end{equation}
\end{proof}
\subsection{Proof of Proposition 1}
\begin{proof}
The Gaussian distribution of the random variable $t$ is expressed as:
\begin{equation}
    \label{eq:equ19}
    f(t)=\frac{1}{\sqrt{2\pi}\Sigma}e^{\{-\frac{(t-\mu)^2}{2\Sigma^2}\}}.
    \end{equation}
According to the definition in Equation~(\ref{eq:equ3}), the first characteristic function of random variable $t$ is calculated as:\\
\begin{equation}
    \label{eq:equ20}
    \begin{aligned}
      &\phi(s)=\int_ {-\infty}^{+\infty} \frac{1}{\sqrt{2\pi}\Sigma}e^{\{-\frac{(t-\mu)^2}{2\Sigma^2}\}}e^{st}dt\\
      &\xlongequal[]{t'=t-\mu}\int_ {-\infty}^{+\infty} \frac{1}{\sqrt{2\pi}\Sigma}e^{\{-\frac{(t')^2}{2\Sigma^2}\}}e^{s(t'+\mu)}dt'\\
      &=e^{\mu s}\int_ {-\infty}^{+\infty} \frac{1}{\sqrt{2\pi}\Sigma}e^{\{-\frac{(t')^2}{2\Sigma^2}\}}e^{st'}dt'\\
      &=e^{\mu s}\frac{1}{\sqrt{2\pi}\Sigma}\int_ {-\infty}^{+\infty} e^{\{-\frac{t'^2}{2\Sigma^2}+st'\}}dt'.
    \end{aligned}
    \end{equation}
The common Gaussian integral formula is expressed as:
\begin{equation}
    \label{eq:equ21}
    \begin{aligned}
      \int_ {-\infty}^{+\infty} e^{(-Ax^2\pm2Bx-C)}dx=\sqrt{\frac{\pi}{A}e^{(-\frac{AC-B^2}{A})}}.
    \end{aligned}
    \end{equation}
In the right side of Equation~(\ref{eq:equ21}), let $A=\frac{1}{2\Sigma^2}$, $B=s/2$, $C=0$, the Euqation~(\ref{eq:equ20}) becomes:
\begin{equation}
    \label{eq:equ22}
    \begin{aligned}
      &\phi(s)=e^{\mu s}e^{\frac{1}{2}\Sigma^2s^2}.
    \end{aligned}
    \end{equation}
The second characteristic function of $\psi(s)$ is formulated as:
\begin{equation}
    \label{eq:equ23}
    \begin{aligned}
      \psi(s)=ln\phi(s)=ln(e^{\mu s}e^{\frac{1}{2}\Sigma^2s^2 s})=\mu s+\frac{1}{2}\Sigma^2 s^2.
    \end{aligned}
    \end{equation}
Compare Equation~(\ref{eq:equ23}) with Equation~(\ref{eq:equ5}), the following coefficients of the term $s^o$ can be obtained in Equation~(\ref{eq:equ5}):
\begin{equation}
    \label{eq:equ22}
    \begin{aligned}
      c_1=\mu, c_2=\Sigma^2, c_o=0\quad(o=3,4,...).
    \end{aligned}
    \end{equation}

\end{proof}
\subsection{More Experiments}
\begin{figure}[ht]
\hspace{-0.2cm}
	\subfigure[5-way 1-shot]{
	\begin{minipage}[t]{0.5\linewidth}
    \includegraphics[width=1.7in]{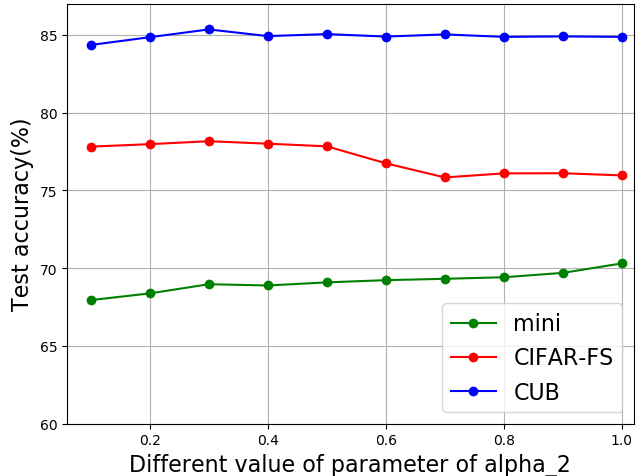}
    \end{minipage}%
     }%
    \subfigure[5-way 5-shot]{
	\begin{minipage}[t]{0.5\linewidth}
    \includegraphics[width=1.7in]{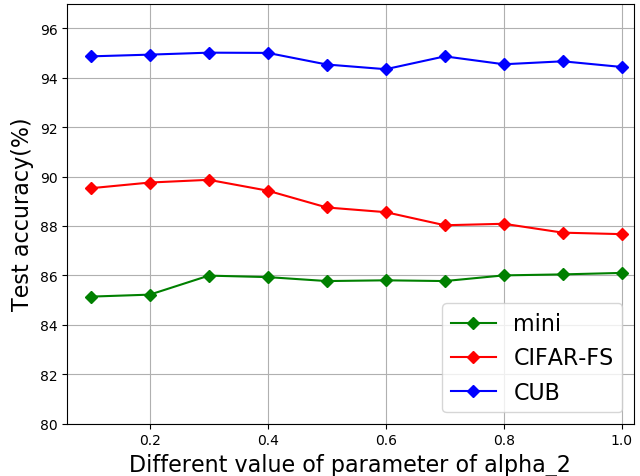}
    \end{minipage}%
     }%
\caption{Test accuracy (\%) under different values of the parameter $\alpha_2$ in the setting of 5-way 1-shot and 5-shot on three FSC datasets.}
\label{fig:Parameter1}
\end{figure}
\begin{figure}[ht]
\hspace{-0.2cm}
	\subfigure[5-way 1-shot]{
	\begin{minipage}[t]{0.5\linewidth}
    \includegraphics[width=1.7in]{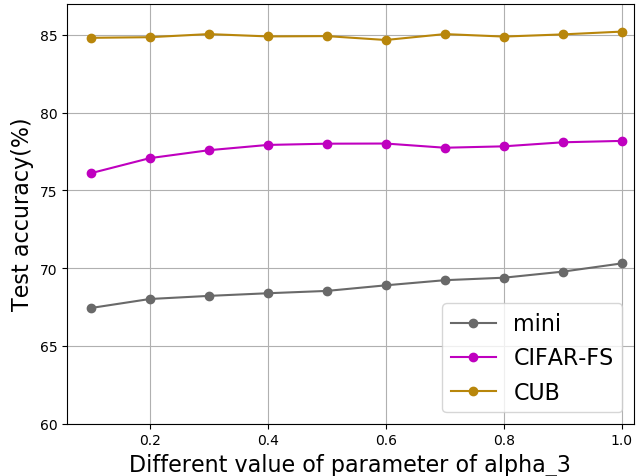}
    \end{minipage}%
     }%
    \subfigure[5-way 5-shot]{
	\begin{minipage}[t]{0.5\linewidth}
    \includegraphics[width=1.7in]{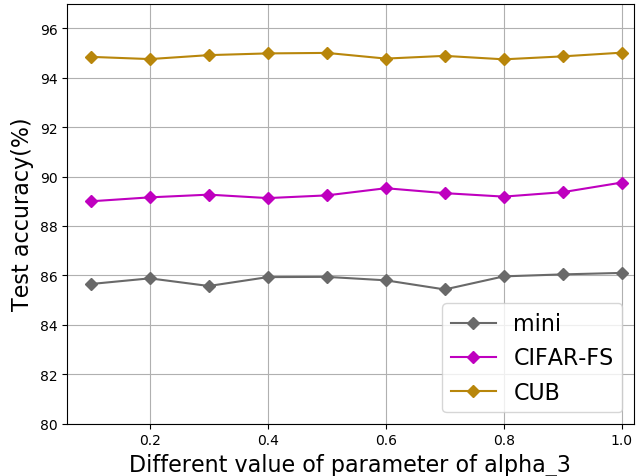}
    \end{minipage}%
     }%
\caption{Test accuracy (\%) under different values of the parameter $\alpha_3$ in the setting of 5-way 1-shot and 5-shot on three FSC datasets.}
\label{fig:Parameter2}
\end{figure}
\begin{figure}[ht]
	\vspace{-0.1cm}
			\includegraphics[width=1\linewidth]{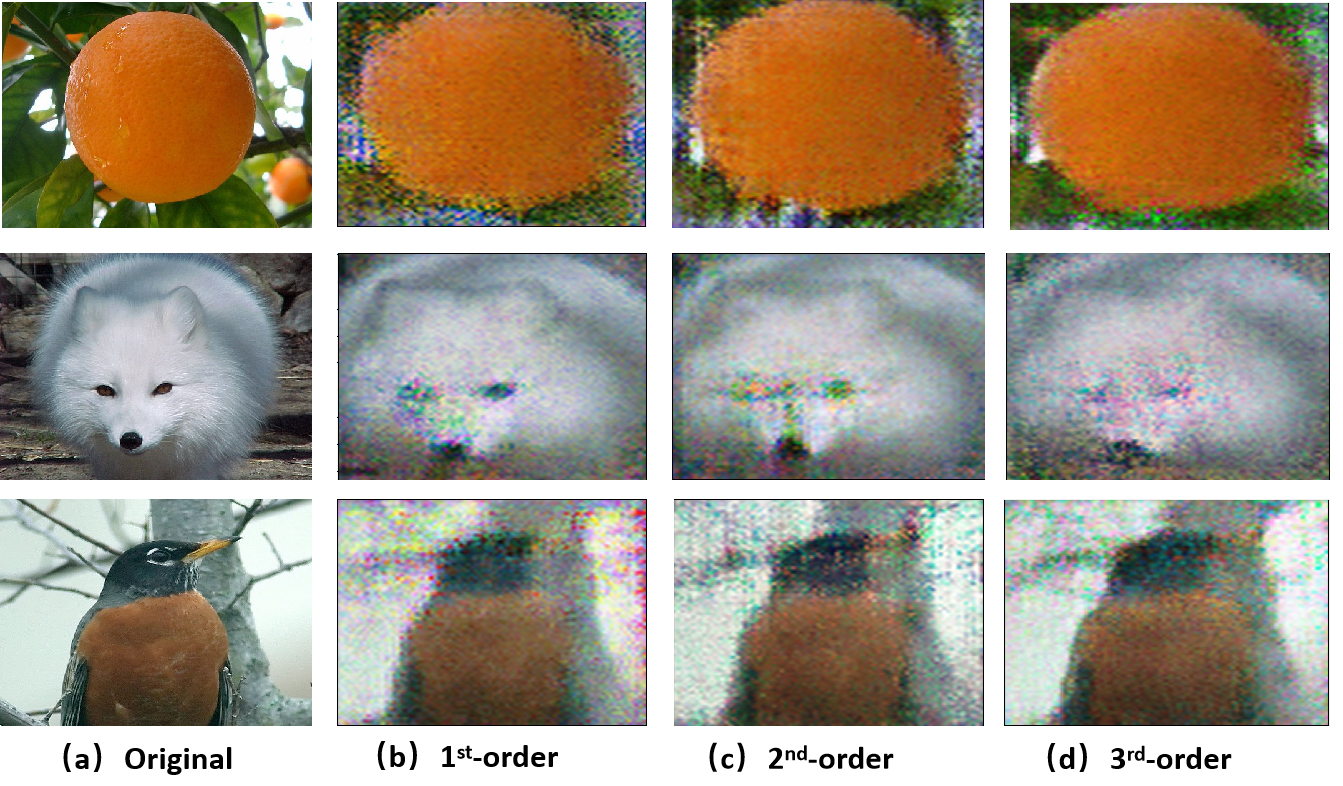}
		\caption{Image reconstruction of features respectively represented by $1^{st}$-order, $2^{nd}$-order, $3^{rd}$-order statistics.}
	\label{fig:reconstruction}
\end{figure}
\begin{figure}[ht]
\hspace{-0.2cm}
	\subfigure[Baseline]{
	\begin{minipage}[t]{0.5\linewidth}
    \includegraphics[width=1.7in]{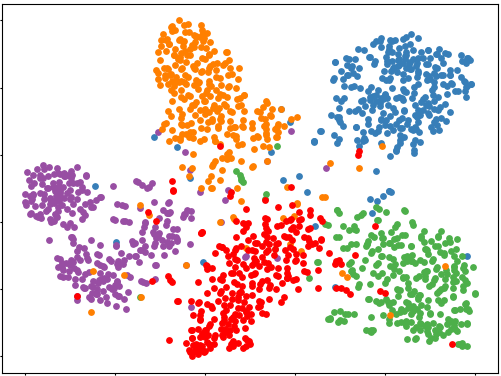}
    \end{minipage}%
     }%
    \subfigure[Our method]{
	\begin{minipage}[t]{0.5\linewidth}
    \includegraphics[width=1.7in]{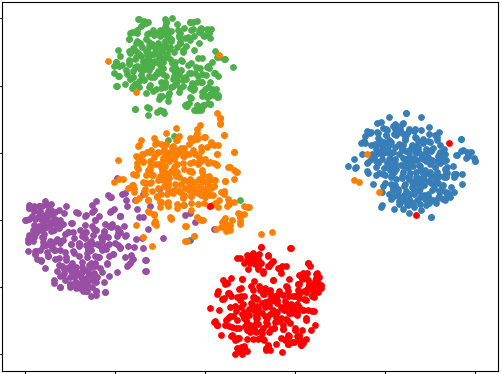}
    \end{minipage}%
     }%
\caption{T-SNE visualization of features on unseen samples of Baseline and our method. }
\label{fig:tsne}
\end{figure}
\begin{table}[ht]
    \renewcommand\tabcolsep{1.5pt} 
    \small
    \caption{Comparison of different methods under cross-domain scenario.}
    \label{tbl:Table4}
    \begin{center}
    \begin{tabular}{p{5cm}ccc}
	\hline
	\multirow{2}{*}{Method} & \multicolumn{2}{c}{miniImageNet $\rightarrow$ CUB}
	 
	\\ \cline{2-3}  &1-shot &5-shot  \\ \hline
	Prototypical~\cite{snell2017prototypical}†&36.61$\pm$0.53&55.23$\pm$0.83 \\
	Relational~\cite{sung2018learning}†&44.07$\pm$0.77&59.46$\pm$0.71\\
	MetaOptNet~\cite{bertinetto2018meta}††&44.79$\pm$0.75&64.98$\pm$0.68\\
	IEPT~\cite{zhang2020iept}&\textcolor{blue}{52.68$\pm$0.56}&\textcolor{blue}{72.98 $\pm$0.40 }\\
	FPN~\cite{wertheimer2021few}&\textcolor{green}{51.60$\pm$0.21}&\textcolor{green}{72.97$\pm$0.18}\\
	BML~\cite{zhou2021binocular}&-&72.42$\pm$0.54 \\ 
	Baseline++~\cite{chen2019closer}&-&62.04$\pm$0.76\\
	SimpleShot~\cite{wang2019simpleshot}††&48.56& 65.63\\
	S2M2~\cite{mangla2020charting}&48.24$\pm$0.84 &70.44$\pm$0.75\\
	Neg-Cosine~\cite{liu2020negative}&-&67.03$\pm$0.76\\
    GNN+FT~\cite{tseng2020cross}&47.47$\pm$0.75&66.98$\pm$0.68\\
	ELMOS(ours)&\textcolor{red}{53.73$\pm$0.47}&\textcolor{red}{74.37$\pm$0.37} \\
	\hline
    \end{tabular}
    \end{center}
\end{table}
\subsubsection{Parameter Analysis}
The effect of each branch is controlled by the parameters $\alpha_1$, $\alpha_2$ and $\alpha_3$ in Equation~(\ref{eq:equ10}). Since the first branch modeling the $1^{st}$-order statistic is the main branch, we set its corresponding parameter to  1. Subsequently, we first fixed the value of $\alpha_3$ to be 1, and varied the value of $\alpha_2$ between [0, 1] with an interval of 0.1. The test accuracy under different values is shown in Figure~\ref{fig:Parameter1}. When $\alpha_2$ is 1, the highest performance on miniImageNet under both 1-shot and 5-shot tasks can be achieved. When $\alpha_2$ is 0.3, we get the highest performance on CUB and CIFAR-FS under both 1-shot and 5-shot tasks. Next, we fixed the value of $\alpha_2$ to be 1 on miniImageNet, 0.3 on CIFAR-FS and CUB, and varied the value of $\alpha_3$ between [0, 1] with an interval of 0.1. The test accuracy under different values is shown in Figure~\ref{fig:Parameter2}. When $\alpha_3$ is 1, we get the highest performance on all three datasets under both 1-shot and 5-shot tasks.

\subsubsection{Image Reconstruction of Features}
The effectiveness of our method is mainly attributed to the diversity of $1^{st}$-order, $2^{nd}$-order, and $3^{rd}$-order statistic features. We used the technique of deep image prior to respectively invert different-order statistic features after the pre-training into RGB images. The reconstruction results are shown in Figure~\ref{fig:reconstruction}. From the results, we notice that as the order of statistic feature becomes higher, the reconstructed images become more smooth. The above phenomenon illustrates that $2^{nd}$-order and $3^{rd}$-order statistic features are more robust to the singularity variation such as the noise point than the $1^{st}$-order statistic feature. By comparison, the $1^{st}$-order statistic feature has stronger ability of capturing the details of the images than $2^{nd}$-order and $3^{rd}$-order statistic features. The above analysis has shown that $1^{st}$-order, $2^{nd}$-order, and $3^{rd}$-order statistic features are  complementary.
\subsubsection{t-SNE Visualization of Features}
 To show the performance of our method, we visualize the features of the novel class samples in comparison with the Baseline. Herein, the Baseline pre-trained the backbone network only with the global average pooling. We randomly selected 5 classes and 200 samples per class from CIFAR-FS and visualize the features of the samples using t-SNE. The visualization results are shown in Figure~\ref{fig:tsne}. From the results, we can see that five classes can well separate from each other in our feature space compared to with the Baseline, which illustrates that our method can extract better features for unseen novel classes compared with the Baseline.

\subsubsection{Comparison of Cross-domain Performance}
As stated in our Theorem~\ref{thm-1}, there exists a domain shift between the base and novel classes. In the former test, the base and novel classes are in the same domain, which has a smaller domain divergence than the ones in the different domains. Now, we large the domain divergence to evaluate our method by doing cross-domain FSC. Following the protocol in~\cite{chen2019closer}, the model was trained on miniImageNet and then evaluated on the novel classes in CUB. The comparison of results is shown in Table~\ref{tbl:Table4}. From the results, we can see that our method is better than all the compared methods under 1-shot and 5-shot tasks. Specifically, our method outperforms the best method of IEPT with the improvement of 1.05\% and 1.39\% respectively. Our method does not concern the domain divergence, but we can also get good cross-domain performance by implementing ensemble learning to decrease the generalization error on base classes, because it is also an important term for the true error on novel classes. 


\end{document}